\documentclass[letterpaper, 10 pt, conference]{ieeeconf}  % Comment this line out if you need a4paper

\IEEEoverridecommandlockouts                              % This command is only needed if 
\usepackage{xcolor}
\usepackage{graphicx} % for pdf, bitmapped graphics files

\usepackage{graphicx}
\usepackage{booktabs}
\usepackage{caption}
\usepackage{subcaption}
\usepackage{multirow}
\usepackage{chngpage}
\usepackage{wrapfig}
\usepackage{adjustbox}
\usepackage{comment}
\usepackage{amsmath}
\usepackage{amssymb}
\usepackage{color}
\usepackage{listings}
\usepackage{xcolor}
\definecolor{codegreen}{rgb}{0,0.6,0}
\definecolor{codegray}{rgb}{0.5,0.5,0.5}
\definecolor{codepurple}{rgb}{0.58,0,0.82}
\definecolor{backcolour}{rgb}{0.95,0.95,0.92}

\lstdefinestyle{mystyle}{
    backgroundcolor=\color{backcolour},   
    commentstyle=\color{codegreen},
    keywordstyle=\color{magenta},
    numberstyle=\tiny\color{codegray},
    stringstyle=\color{codepurple},
    basicstyle=\ttfamily\scriptsize,
    breakatwhitespace=false,         
    breaklines=true,                 
    captionpos=b,                    
    keepspaces=true,                 
    showspaces=false,                
    showstringspaces=false,
    showtabs=false,                  
    tabsize=2
}

\usepackage{url}

\usepackage{color}
\usepackage{hyperref}
\usepackage{xspace}

\title{\LARGE \bf
Gen2Sim: Scaling up Robot Learning in Simulation \\ with Generative Models}

\author{Pushkal Katara$^{*}$, Zhou Xian$^{*}$,  Katerina Fragkiadaki% <-this % stops a space
\vspace{6pt}
\\Carnegie Mellon University
\vspace{6pt}
\\ \large{\url{https://gen2sim.github.io/}}
\thanks{$^{*}$ Equal contribution.}
\thanks{The authors are with School of Computer Science, Carnegie Mellon University
        {\tt\small <pkatara, xianz1, katef>@cs.cmu.edu}}%
}

\newcommand{\model}{Gen2Sim}

\newcommand{\bzero}{\mathbf{0}}
\newcommand{\bone}{\mathbf{1}}
\newcommand{\cond}{\textbf{c}}

\begin{document}

\maketitle

\begin{abstract}
%While robot locomotion 
Generalist robot manipulators  need to learn a wide variety of  manipulation skills across diverse environments. Current robot training pipelines rely on humans to provide kinesthetic demonstrations or to program simulation environments and to code up reward functions for reinforcement learning. Such human involvement is an important bottleneck towards scaling up robot learning across diverse tasks and environments. %How can we train robots to master diverse tasks and environments w
We propose \textit{Gen}eration to \textit{Sim}ulation (\textit{\model{}}), a method for scaling up robot skill learning in simulation by automating generation of  3D assets,  task descriptions, task decompositions and reward functions using large pre-trained generative models of language and vision. We generate 3D  assets for simulation by lifting open-world 2D object-centric images to 3D using image diffusion models and querying LLMs to determine plausible  physics parameters. %We generate task descriptions, their decompositions and reward functions by one-shot chain-of-thought prompting LLMs to parse 
Given URDF files of generated and human-developed assets, we chain-of-thought prompt LLMs to map these to relevant  task descriptions, temporal decompositions, and corresponding python reward functions for reinforcement learning.  %, based on the assets and scene affordances.  
%We train  policies with reinforcement learning for the generated task descriptions  using the generated reward functions.
We show  \model{}  succeeds in learning policies  for diverse long horizon tasks,  where reinforcement learning with non temporally decomposed reward functions fails. \model{} provides a viable path for scaling up reinforcement learning for robot manipulators in simulation, both by diversifying and expanding task and environment development, and by facilitating the discovery of reinforcement-learned behaviors through temporal task decomposition in RL. 
% using LLMs' common sense. 
%which would not have been possible without any human involvement. Policy We show construction of simulated twin environment based on real world observations, and successful transfer of the trained robot trajectories from simulation to the real world. 
Our work contributes hundreds of simulated assets, tasks and demonstrations, taking a step towards fully autonomous robotic manipulation skill acquisition in simulation. 
\end{abstract}

\section{Introduction}

Scaling up training data has been a driving force behind the recent revolutions in language modeling \cite{https://doi.org/10.48550/arxiv.2005.14165}, image understanding \cite{CLIP}, speech recognition \cite{radford2022robust}, image generation \cite{rombach2022high}, to name a few.   %Yet, in robotics the datasets remain
%, resulting in large-scale pretrained models, termed foundation, to u.  
This begs the question: can we scale up robot data to enable a similar revolution in robotic skill learning? One way to scale robot data is in the real world, by having multiple robots explore \cite{levine2016learning} or by having humans provide  kinesthetic demonstrations 
%, or collecting those with appropriate instrumentation \cite{}. at scale, with proper instrumentation or crowd-sourcing 
\cite{sieb2020graph, brohan2023rt, brohan2022rt}. %\todo{xian?}
This is a  promising direction; however, safety concerns and wear and tear of the robots hinder robot exploration in the real-world, and  collecting kinesthetic demonstrations scales poorly as it is time-consuming and labor-intensive \cite{brohan2022rt}. 
% For example RT-1 \cite{brohan2022rt} authors spent a year and a half collecting 130K demonstrations. 
%is a laborious  or deployment of half-trained policies. % not permit collecting the millions of 
%does not permit exploration due to safety co 
Another way to scale robot data is in simulation, by developing simulated environments, defining tasks  and their reward functions, and training robot policies with reinforcement learning, augmenting visuals and physics parameters to facilitate   transfer of policies to the real world \cite{DBLP:journals/tase/HoferBHGMGAFGLL21}. %Though simulators have been criticized for being impossible to simulate fine contacts, the community has been exploring hybrid solutions of explicit physics and learning-based residuals.  Learning in simulation and transfering in the real world has had limited success so far. Notable exceptions % computer vision and 
%\topic{Improving simulators is important}
%Improving and scaling up simulators is crucial for robot learning in simulation and zero shot or few shot tranfering in the real world.
%The safety concerns and wear and tear of robots has made sim2real pathway of policy training in simulation and zero-shot or few-shot exploration/deployment a viable soltuion. 
Such sim2real paradigm has seen  recent
successes 
in  robot locomotion \cite{fu2022coupling, kumar2021rma,DBLP:journals/corr/abs-2010-11251,ji2023dribblebot},   object re-orientation   \cite{chen2022system,openai2019solving},  and drone flight \cite{drones}. %, fruit cutting and deformable object manipulation \cite{lin2022diffskill, xu2023roboninja}
These examples,  though very important and exciting, are still fairly isolated. % and focus on learning the robot's dynamics as opposed to modelling its environment. 
%in tasks including robotic locomotion \cite{fu2022coupling, kumar2021rma, radosavovic2023learning, wang2023softzoo}, multi-finger manipulation \cite{chen2022system, chen2022towards}, furniture assembly \cite{suarez2018can}, deformable object manipulation \cite{lin2022diffskill, huang2021plasticinelab, xu2023roboninja, xian2023fluidlab}, acrobatic flight \cite{kaufmann2020deep, loquercio2021learning}. 
%However, these examples are still fairly isolated in task-specific domains with limited generality.
 
A central bottleneck towards scaling up simulation environments and tasks is the laborious manual effort needed for developing the visuals and physics of assets, their spatial arrangement and configurations, the development of task definition and reward functions, or the collection of programmatic demonstrations. 
% These  tasks that requires human expertise, and cannot be crowdsourced. % for robots to solve the tasks. % with reinforcement learning or imitation learning. 
%For sim2real to contribute to large scale robot learning a major bottleneck is simulator's development. 
%Designing 3D assets and their physics parameters, and coding up their fine-grain contact interactions 
% These are laborious tasks that requires human expertise, and cannot be crowdsourced. 
% Many major companies have invested 
Tremendous resources have been invested in developing simulators for autonomous vehicles \cite{dosovitskiy2017carla}, warehouse robots, articulated objects \cite{xiang2020sapien}, home environments \cite{srivastava2021behavior,savva2019habitat,gan2021threedworld}, etc., many of which are proprietary and not open-sourced.  
%Many simulation environments have abstractions of physics, not suitable visuals, or poorly fit the application domain in hand.  
Given these considerations, an important question naturally arises: How can we minimize manual effort in simulation development for diverse robotic skill learning?

%, which often outsource the developed environments . 
%Still, sim2real is far from solved yet. Recent work shows that learning in simulation does not transfer in the real world, but also, policy performance in simulation does not correlate with the real world. This means that potentially.

\begin{figure}[!t]
    \includegraphics[scale=0.3,width=\columnwidth,keepaspectratio]{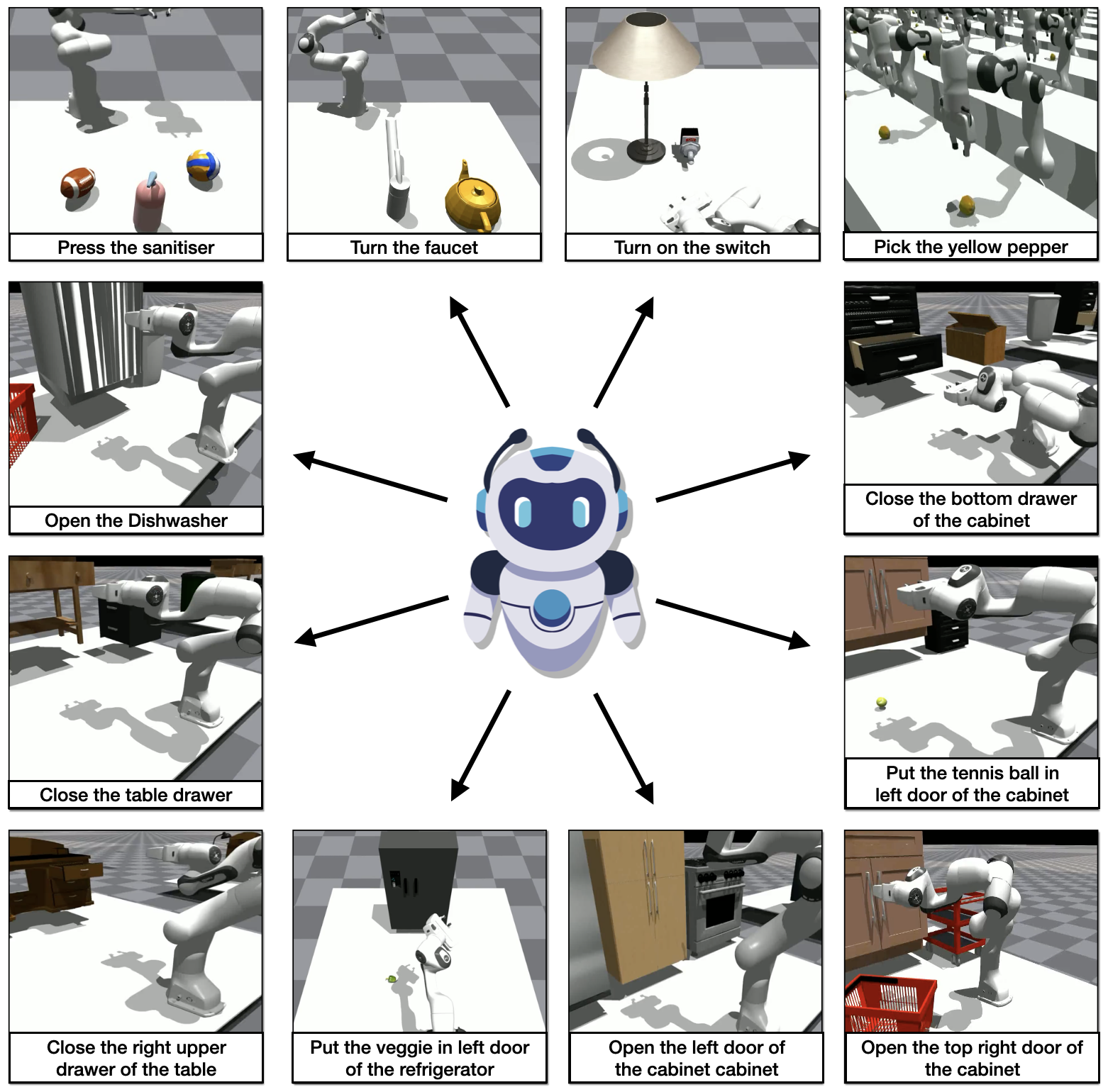}
    \caption{\textbf{\model{}} is an automated generative pipeline of assets,  tasks, task decompositions and  rewards functions for autonomous robotic skill reinforcement learning in simulation. Here we show 12 generated tasks, concerning  affordances of diverse types of object assets and their combinations.}
    \label{fig:teaser}
    \vspace{-0.3in}
\end{figure}

%\topic{what we propose}
In this paper, we explore automating  the development of simulation environments, manipulation tasks and rewards for robot skill learning, by building upon  latest advances in large pre-trained generative models of images and language. Our system strives to automate all stages of robot learning: from generating 3D assets, textures, and physics parameters, to generating task descriptions and reward functions, leading to automated skill learning in diverse scenarios, as shown in Figure \ref{fig:teaser}. This generative pipeline was first proposed in a recent position paper \cite{xian2023towards}, described as a promising pathway towards generating diverse data for generalist robot learning. In this paper, we present \model{}, the first attempt and realization of such a generative paradigm. We automate 3D object asset generation by combining image diffusion models for 3D mesh and texture generation, and LLMs for querying physical parameters information. We showcase how LLMs and image generative models can diversify the appearances and behaviors of assets by producing plausible ranges of textures, sizes and physical parameters, achieving ``intelligent" domain diversification. 
We automate task description, task decomposition and reward function generation by few-shot prompting of LLMs to generate language descriptions for semantically meaningful tasks, concerning affordances of existing and generated 3D assets, articulated or not, alongside their reward functions. \model{} is able to generate numerous object assets and task variations without any human involvement beyond few LLM prompt designs. We successfully train RL policies using our auto-generated tasks and reward functions. We also demonstrate the usefulness of our simulation-trained policies, by constructing  digital-twin environments from given real scenes, allowing a robot to practice skills in the twin simulator and deploying it back to the real world to execute the task.

In summary, we make the following contributions:
\begin{itemize}
    \item We show how pre-trained generative models of images and language can help automate 3D asset generation and diversification, task description generation, task decomposition and reward function generation that supports reinforcement learning of long horizon tasks in simulation with minimal human involvement.  
    \item We deploy our method to generate hundreds of assets, and hundreds of manipulation tasks, their decompositions   and  their reward functions, for both human-developed and  automatically generated object assets. 
\end{itemize}
For code, videos and  qualitative video results, please visit our project website: \url{https://gen2sim.github.io/}.

% \begin{figure}
%     \centering
%     \includegraphics[height=9cm,keepaspectratio]{draft_figures/tasks}
%     \caption{\textbf{Overview for \model{}.} \model{} explores how to automate the pipeline of generating simulation environments, tasks and reward functions  for robot manipulation skill learning by building upon recent progress on large scale generative models of language and vision. It creates 3D models of objects from individual, real or generated images using image diffusion models, assigns plausible ranges for 3D sizes and physics parameters using LLMs and introduces the generated assets into a physics engine. It then prompts LLMs to suggest relevant task and reward functions per asset or scene, based on object and scene affordances. It uses reinforcement learning to solve the generated tasks using the generated reward function in the generated environments.  }
%     \label{fig:architecture}
% \end{figure}

%\input{2_related_work}
\begin{figure*}[!ht]
     \centering
     \includegraphics[width=\textwidth]{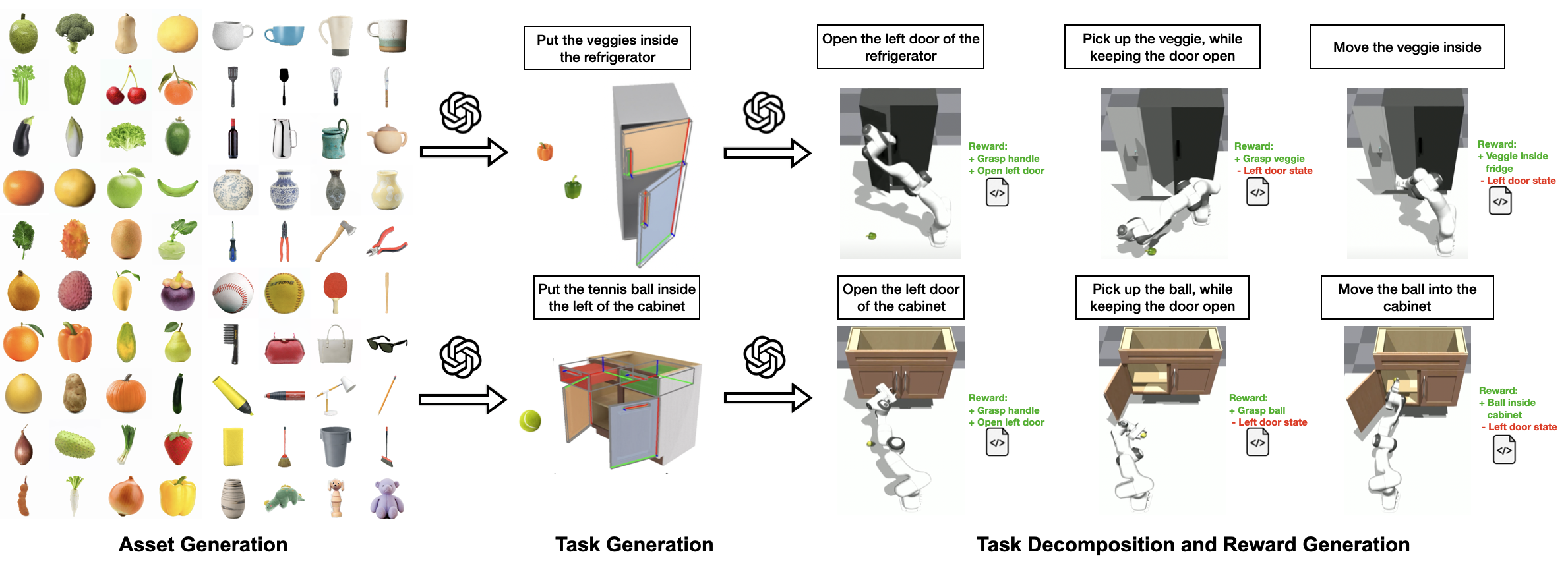}
    \caption{ \textbf{The \model{} components.} \model{} generates  3d assets by lifting object-centric 2D images to 3D. It then uses both generated assets and assets obtained from other publicly available datasets to populate scene environments. Afterwards, it queries LLMs to generate meaningful task descriptions for the assembled scenes, as well as  decompose the generated task descriptions to sub-tasks and their reward functions.}
     \label{fig:method}
    % \vspace{-0.2in}
\end{figure*}

%\todo{pushkal to add the image paired with 3d }
\section{Related Work}

%\vspace{-0.05in}
\textbf{Large Language Models for task and motion planning in robotics}
Large language models (LLMs)  map instructions to language subgoals \cite{DBLP:journals/corr/abs-2012-07277,DBLP:journals/corr/abs-1909-13072,huang2022language,https://doi.org/10.48550/arxiv.2207.05608} or action programs \cite{LMP} with appropriate plan-like or program-like prompts.
LLMs trained from Internet-scale text have shown impressive zero-shot reasoning capabilities for a variety of downstream language tasks \cite{https://doi.org/10.48550/arxiv.2005.14165} when prompted appropriately, without any weight fine-tuning \cite{DBLP:journals/corr/abs-2201-11903,DBLP:journals/corr/abs-2107-13586,brown2020language,chan2022data}. 
LLMs were used to generate task curricula and predict skills to execute in Minecraft worlds \cite{zhu2023ghost,lifshitz2023steve1,wang2023voyager}
%, map instructions to language subgoals  \cite{DBLP:journals/corr/abs-2012-07277,DBLP:journals/corr/abs-1909-13072,huang2022language,https://doi.org/10.48550/arxiv.2207.05608} or program-like policies \cite{LMP}.  %Methods of \cite{wang2023voyager} use privileged information from the Minecraft simulator's interface regarding failures, state and inventory which they use to feedback to an LLM to plan and re-plan based on sub-goal successes and failures. 
%LLMs are also being used as knowledge sources in place of traditional triplet structured knowledge bases  \cite{roberts2020knowledge,brown2020language}. 
Following the seminal work of Code as Policies, many works map language to programs over given skills \cite{ha2023scaling} or hand-designed motion planners \cite{huang2023voxposer}. 
Our work instead maps task descriptions into task decompositions and reward functions, to guide reinforcement learning in simulation, to discover behaviours that  achieve the generated tasks. Work of \cite{yu2023language} also uses language for predicting reward functions for robot locomotion, but does not consider task generation and decomposition or interaction with objects. Our work is the first to use LLMs for task decomposition and reward generation, as well as asset generation.   

%\todo{we need to mention deepmind language to reward paper}

% can learn open a door
%\vspace{-0.05in}
% \paragraph{Automating 3D asset creation with  generative models and differentiable simulators}
\textbf{Automating 3D asset creation with generative models} 
%From real fusion
%\todo{some philosophy about image to simulation in the brain}
The traditional process of creating 3D assets typically
involves multiple labor-intensive stages, including geometry modeling, shape baking, UV mapping, material creation,  texturing and physics parameter estimation, where different software
tools and the expertise of skilled artists are often required. 
%Imperfections would also accumulate across these stages, resulting in low-quality 3D assets.
It is thus desirable to automate 3D asset generation to automatically generate high-quality assets that support realistic rendering  under arbitrary views and have plausible physical behaviours during force application and contacts. 
The lack of available 3D data and the abundance of 2D image data have stimulated interest in learning 3D  models from  2D image generators \cite{nguyenphuoc2019hologan, chan2021pi}.  
%The alternative to training a 3D diffusion model is to extract 3D information from an existing 2D model. 
%A 2D image generator can  be used to sample or validate multiple views of a given object; these multiple views can then be used to perform 3D reconstruction. Early GAN-based generators led to some success for simple data like faces and synthetic objects \cite{nguyenphuoc2020blockgan,nguyenphuoc2019hologan,chan2022efficient}.  
%With the availability of large-scale models like CLIP [34]and, more recently, diffusion models, increasingly complex results have been obtained. 
The availability of strong 2D image generative models based on diffusion led to high-quality 3D models from text descriptions \cite{poole2022dreamfusion,chen2023fantasia3d,lin2023magic3d} or single 2D images using the diffusion model as a 2D prior \cite{melaskyriazi2023realfusion,tang2023makeit3d,shen2023gina3d}.  
In this work, instead of a text-conditioned model, we use a view and relative pose conditioned image generative model, which we found to provide better prior for score distillation. 
Some methods attempt to use videos of assets and differentiable simulations to estimate their physics parameters and/or adapt the simulation environment, in an attempt to close the simulation to reality gap \cite{heiden2022probabilistic,heiden2022disect,wang2023real2sim2real}. 
%\todo{waymo pap[er + any other recent papers]}
Our effort is complementary to these works.
%We opt for exploiting knowledge of real-world distributions encoded in LLMs for physics value ranges. 
% We see sim2real2sim2real and a very plausible pipeline for few shot adaptation of simulation environments. 

%\todo{to mention behaviour}
%\vspace{-0.05in}
%\todo{we can generate a language goal or image goal also}
\textbf{Procedural demonstration generation using symbolic planners} 
Many recent works procedurally generate  scenes and demonstration trajectories using planners that have access to privileged information to solve the task, and distill the demonstration solutions into learning-based policies that operate directly from pixel or point-cloud input \cite{fishman2022motion,dalal2023imitating,mcdonald2021guided}. 
%For example, motion policy networks \cite{fishman2022motion} train neural networks to predict obstacle avoiding end-effector  trajectories in free space by imitating the solutions of RRT planners.
% CabiNet \cite{murali2023cabinet} trains neural networks for collision detection by imitating ground-truth collision information by placing objects in different locations in the robot's space. 
%The learning 0-based policies are faster at inference time and have only access to available sensory input.   follows a si
%Dieter in motion policy nets
%imitating tamp
%CabiNet- Scaling Neural Collision Detection for Object Rearrangement with Procedural Scene Generation
%Guided imitation of task and motion planning.
%Works of \cite{dalal2023imitating,mcdonald2021guided} train pick and place policies by  imitating task and motion planners (TAMP) in procedurally generated scenes. 
%\textit{Symbolic planners}  such as  PDDL (Planning Domain Definition Language)   planners 
Task and motion planners 
\cite{DBLP:journals/corr/abs-1911-04679,5980391, 10.5555/2832415.2832517,DBLP:journals/corr/abs-1811-00090} use predefined symbolic rules and known dynamics models, and infer discrete task plans given instruction with lookahead logic search \cite{5980391,DBLP:journals/corr/abs-1802-08705,DBLP:journals/corr/abs-1911-04679,5980391, 10.5555/2832415.2832517,DBLP:journals/corr/abs-1811-00090}. 
%Symbolic planners  assume that each state of the world,  scene goal and intermediate subgoal  can be sufficiently represented in a logical form, using language predicates that describe object spatial relations. 
These methods predominantly rely on manually-specified symbolic transition rules, planning domains, and grounding, which limits their applicability. 
%TAMP systems that operate over discrete symbol transition functions and use graph search to find solutions \cite{}. TAMP and motion planner can mainly find solutions to pick and place symbolically defined tasks, and have limitation into discovering skills for opening a door. 
Indeed, works of \cite{dalal2023imitating,mcdonald2021guided}   demonstrate their results on relatively simple multi-object box stacking tasks. 
Scene procedural generation in the aforementioned works \cite{dalal2023imitating,mcdonald2021guided,murali2023cabinet} entails randomizing locations and number of given 3D models under weak supervision from a human that defines the task and the possible location candidates.
%a human defining the type of objects and environment, e.g.,  moving an object from the lower rung of the shelf to the middle one, and objects arrangements are randomized. Solutions for the task are provided by TAMP systems that operate over discrete symbol transition functions and use graph search to find solutions. They have been demonstrated mainly on tasks related to multi-object box stacking. TAMP and motion planner can mainly find solutions to pick and place symbolically defined tasks, and have limitation into discovering skills for opening a door.
In contrast, we unleash the common sense knowledge and reasoning capabilities provided by LLMs and use them to suggest task descriptions, task decompositions, and reward functions. We then use reinforcement learning to discover solution trajectories instead of TAMP-based search.  
% We are able to address tasks that TAMP systems do not know how to solve, such as manipulating articulated objects. %LLMs generalize over TAMP  on their wide knowledge of how the world works, and have shown % from the internet. 
%We are thus able to automate environment and task generation for home and food related assets we create, which was not possible with existing scene procedural generation works.
%not available on online 3D object repositories \cite{}. We believe our method goes well beyond what was possible with previous procedural generation works. % We train with RL in simulation and imitate with a vision-based policy, similar to previous works. 

\textbf{Simulation environments for robotic skill learning}
In recent years, improving simulators for robot manipulation has attracted increasingly more attention. 
Many robotic manipulation environments and benchmarks \cite{brockman2016openai, kolve2017ai2, xiang2020sapien} are built on top of either PyBullet \cite{coumans2019} or MuJoCo \cite{todorov2012mujoco} as their underlying physics engines, which mainly support rigid-body manipulation \cite{tung20203d, gkanatsios2023energy, xian2021hyperdynamics, gervet2023act3d, xian2023chain}. 
Recently, environments supporting soft-body manipulation (\cite{macklin2014unified, xiang2020sapien, gan2020threedworld, lin2020softgym, xian2023fluidlab, wang2023softzoo}) provide capabilities for simulating deformable robots, objects and fluids.  
Our automated asset and task generation are not tied to any specific simulation platforms and can be used with any of them.

\section{\model{}}
% \todo{The overview for \model{} is shown in Figure \ref{fig:architecture}.} 
\model{} generates 3D assets from object-centric images using image diffusion models and predicts physical parameters for them using LLMs (Section \ref{sec:assets}). It then prompts LLMs to generate language task descriptions and corresponding reward functions for each generated or human-developed asset,  suitable to their affordances (Section \ref{sec:prompting}). Finally, we train RL policies in the generated environments using the generated reward functions. We additionally show the applicability of the simulation-trained policy by constructing digital twin environment in simulation, and deploy the trained trajectory in the real world (Section \ref{sec:RL}). See Figure \ref{fig:method} for our method overview.

%\vspace{-0.05in}
\subsection{3D Asset Generation} \label{sec:assets} % using vision and language generatiuve models}
%Generating appearance, geometry and physics for 3D assets from images or text using image and language large scale pre-trained generative models}
\model{} automates 3D asset generation by %starts the automation pipeline by 
mapping 2D images of objects to textured 3D meshes with plausible physics parameters. 
The  images can be 1) real images taken in the robot's environment, 2) real images provided by Google search under relevant category names, e.g., \textit{``avocado"}, or 3) images generated by  pre-trained text-conditioned diffusion models, such as stable diffusion \cite{rombach2022highresolution}, prompted appropriately to generate uncluttered images of the relevant objects, e.g., \textit{``an image of an individual avocado"}.  We %automate the generation of lists of interesting object categories by 
query GPT-4 \cite{openai2023gpt4} for a list of object categories relevant for manipulation tasks to search online for or to generate, instead of manually designing it. Please, visit our project site for a detailed list of the objects we generated. 
%In this way, our method can support generation of a large    number of assets with minimal to no human involvement. 
%In our appendix, we show a list of category names produced by GPT-4 for Google search and image generation. 
Given a real or generated 2D image of an object, we lift it to a 3D model by minimizing re-reprojection error  and maximizing  likelihood of its image renderings using a diffusion model   \cite{poole2022dreamfusion, chen2023fantasia3d}. %, initially developed in \cite{poole2022dreamfusion, wang2022score} for text-to-3D lifting. 
We provide background on image diffusion models below, 
before we describe our 3D model fitting approach. % in Section \ref{sec:phys}. 
%\ref{sec:SDS}
%generating queries for objects. We feed a simple prompt asking GPT-4 \cite{gpt4} about interesting objects that are commonly seen in robotic manipulation scenario. We show an example of such prompt and returned response from GPT-4 in \xian{Appendix \ref{sec:}}. Afterwards, we feed the generated object queries into text-based diffusion models for 2D image generation \xian{and then multi view generation?}.
%in the supplementary file.
%\xian{why do we need so much details for diffusion? just say what models we use and how we use it...}
%\subsubsection{Background} \label{sec:background}
%\vspace{-0.05in}
\subsubsection{\textbf{Image diffusion models}}
A diffusion model learns to model a probability distribution $p(x)$ by inverting a process that gradually adds noise to the image $x$. 
The diffusion process is associated with a variance schedule $\{\beta_t \in (0,1)\}_{t=1}^{T}$, which defines how much noise is added at each time step. The noisy version of sample $x$ at time $t$ can then be written $x_t = \sqrt{\bar{\alpha}_t} x + \sqrt{1-\bar{\alpha}_t} \epsilon$ where
$
\epsilon \sim \mathcal{N}(\bzero,\bone),
$
is a sample from a Gaussian distribution (with the same dimensionality as $x$), $\alpha_t = 1- \beta_t$, and $\bar{\alpha}_t = \prod_{i=1}^t \alpha_i$.
One then learns a denoising neural network
$
\hat \epsilon = \epsilon_{\phi}(x_t;t)
$
that takes as input the noisy image $x_t$ and the noise level $t$ and tries to predict the noise component $\epsilon$. 
Diffusion models can be easily extended to draw samples from a distribution $p(x|\cond)$ conditioned on a prompt $\cond$, where $\cond$ can be a text description, a camera pose, and image semantic map, {\it etc} ~\cite{rombach2022high,li2023gligen,zhang2023adding}.  
Conditioning on the prompt can be done by adding $\cond$ as an additional input of the network $\epsilon_\phi$.  
For 3D lifting, we build on Zero-1-to-3 \cite{liu2023zero1to3}, a diffusion model for novel object view synthesis that 
conditions on an image view of an object and a relative camera rotation around the object to generate plausible images for the target object viewpoint, $ \cond = [I_1,\pi] $.  
It is trained on a large collection $\mathcal{D'} = \{(x^i, \cond^i)\}_{i=1}^N$ of images paired with views and relative camera orientations as conditioning prompt by minimizing the loss:
\begin{equation*}\label{e:diffloss}
\mathcal{L}_\text{diff}(\phi;\mathcal{D}')
=
\tfrac{1}{|\mathcal{D}'|}
\sum_{x^i, \cond^i\in\mathcal{D}'}
|| \epsilon_{\phi}(\sqrt{\bar{\alpha}_t} x^i + \sqrt{1-\bar{\alpha}_t} \epsilon, \cond^i , t) - \epsilon ||^2.
\end{equation*}
%This loss, which trains the network $\epsilon_\phi$ to predict the noise $\epsilon$ added to an image, corresponds to a reweighted form of the variational lower bound for $\log p(x | \cond)$~\cite{ho2020denoising}. 

%\vspace{-0.05in}
%\textbf{Differentiable 3D model representations: NeRFs and textured meshes:} We employ Neural Radiance Fields (NeRF) \cite{DBLP:journals/corr/abs-2003-08934}  and textured meshes \cite{DBLP:journals/corr/abs-2111-04276} as  our 3D assert representations in a two stage 3D fitting.  NeRFs represent 3D objects using a differential function with parameters $\theta$ that takes coordinates in a 3D space as input and outputs the corresponding color and density. Here, $\theta$ corresponds to the parameters of the function. Given camera pose $\pi$, the rendering process $R(\theta, \pi)$ is defined as casting rays from pixels and computing the weighted sum of the color of the sampling points along each ray to composite the color of each pixel. Textured mesh \cite{DBLP:journals/corr/abs-2111-04276} represents the geometry of a 3D object with triangle meshes and the texture with color on the mesh surface. Here the 3D parameter $\theta$ represent the coordinates of triangle meshes and parameters of the texture. The rendering process $R(\theta, \pi)$  given camera pose $\pi$ is defined by casting rays from pixels and computing the intersections between rays and mesh surfaces to obtain the color of each pixel. 

\subsubsection{\textbf{Image-to-3D Mesh using Score Distillation Sampling}}%\label{sec:SDS}
%From: ProlificDreamer: High-Fidelity and Diverse Text-to-3D Generation with Variational Score Distillation
%SDS is an optimization method by distilling pretrained diffusion models, also known as Score Jacobian Chaining (SJC) [54]. It is widely used in text-to-3D generation [33, 54, 20, 28, 54, 4] with great promise. Given a pretrained text-to-image diffusion model pt(xt|y) with the noise prediction network εpretrain(xt, t, y), SDS optimizes a single 3D representation with parameter θ ∈ Θ, where Θ is the space of θ with the Euclidean metric. Given a camera parameter c with a distribution p(c) and a differentiable rendering mapping g(·,c) : Θ → Rd, denote qtθ(xt|c) as the distribution at time t of the forward diffusion process starting from the rendered image g(θ,c) with the camera c and 3D parameter θ. SDS optimizes the parameter θ by solving min LSDS(θ) := Et,c ��(σt/αt)ω(t)DKL(qtθ(xt|c) ∥ pt(xt|y))�� , (2) θ∈Θ wheret∼U(0.02,0.98),ε∼N(0,I),andxt =αtg(θ,c)+σtε.Itsgradientisapproximatedby �� ∂g(θ,c)�� ∇θLSDS(θ) ≈ Et,ε,c ω(t)(εpretrain(xt, t, y) − ε) ∂θ . (3)  Notwithstanding this progress, empirical observations [33] show that SDS often suffers from over- saturation, over-smoothing, and low-diversity issues, which have yet to be thoroughly explained or adequately addressed.
Given an image and relative camera pose 2D diffusion model $p(I|[I_0,\pi])$,  
we extract from it a 3D rendition of the input image $I_0$, represented by a differential 3D representation using Score Distillation Sampling (SDS) \cite{poole2022dreamfusion, wang2022score}.   
We do so by randomly sampling a camera pose $\pi$, rendering a corresponding
view $I_\pi$, assessing the likelihood of the view based on a diffusion  
model $p(I_\pi|[I_0,\pi])$, and updating the differentiable 3D representation to increase the likelihood of the generated view based on the model. 
%In practice, DreamFusion uses 
Specifically, the diffusion model is frozen and the 3D model is updated as:% a
%frozen critic and takes a gradient step
\begin{equation*}
\begin{aligned}
\nabla(\theta)\mathcal{L}_{SDS}(\theta; \pi, \cond, t) 
 &= \\\mathbb{E}_{t,\epsilon}[w(t)&(\epsilon_\phi(a_t I + \sigma_t\epsilon;t, \cond) - \epsilon) \cdot \nabla_{\theta} I ], 
\end{aligned}
\end{equation*}
where $I = R(\theta, \pi)$ is the image rendered from a given
viewpoint $\pi$. %This process is called Score
%Distillation Sampling (SDS). 
%Note that Eq. (4) differs from simply optimizing the standard diffusion model objective because it does not include the Jacobian term for Φ. In practice, removing thisc term both improves generation quality and reduces computational and memory requirements. 
The loss we use to backpropagate to  the 3D model parameters $\theta$ includes an image re-projection loss for the camera viewpoint of the input image, and score distillation for the other views, using a pre-trained view and pose conditioned image diffusion model of \cite{liu2023zero1to3} to measure 2D image likelihood. 
%In Section \ref{sec:experiments}, we compare our image-to-3D method with RealFusion \cite{} and Fantasia3D \cite{} that use image reprojection losses but evaluate likelihood using text-conditioned generating models as opposed to view and pose conditioned ones, and show the superior performance of our method  %, the total loss reads:
%. We feed the appropriate relative pose change for alternative viewpoints into zero-1-to-3 model. 
We use a two-stage fitting, wherein the first stage an instantNGP NeRF representation \cite{M_ller_2022} is used, similar to RealFusion \cite{melaskyriazi2023realfusion}, and in the second stage a mesh-based representation is initialized from the NeRF and finetuned differentiably, similar to Fantasia3D \cite{chen2023fantasia3d}. More information of our score distillation sampling can be found in our website.  

%SDS was initially developed in \cite{poole2022dreamfusion, wang2022score} for text-to-3D lifting.  \model{} considers an image as input for 3D lifting instead. 
%During NeRF fitting, we use surface normal matching and segmentation mask losses similar to RealFusion. During mesh fitting, we follow Fantasia3D to disentangle the optimization of geometry and texture by first optimizing the geometry using the normal map and then optimizing the texture. 
%We found the image likelihood provided by the view conditioned image generation model to be more useful and informative than a general text-conditioned model. 
 %we compare against RealFusion \cite{melaskyriazi2023realfusion} and Fantasia3D \cite{chen2023fantasia3d} that also consider image-to-3D lifting by textual inversion for diffusion adaptation and by an image re-projection loss, respectively. 

%In Section \ref{sec:experiments}, we show our model generates more faithful 3D models from images because the image likelihood provided by the view and pose conditioned image generative model \cite{tang2023makeit3d} is more informative than a generic or personalized text-conditioned one, used in \cite{melaskyriazi2023realfusion,chen2023fantasia3d}.

\subsubsection{\textbf{Texture generation}} We augment the textures of our generated assets using the method of TEXTure \cite{richardson2023texture} which iteratively edits a mesh's texture by rendering the mesh from different viewpoints and updating the rendered 2D images. While domain randomization \cite{tobin2017domain} randomly re-textures simulated assets, TEXTure produces diverse yet plausible texture augmentations.

%\vspace{-0.05in}
\subsubsection{\textbf{Generating plausible physical properties}}
%\label{sec:phys}
%\vspace{-0.05in}
%We query GPT-4 for 3D sizes and physics parameters 
The visual and collision parameters of an asset are generated from the Image-to-Mesh pipeline discussed above. To define 3D sizes and physics parameters for the generated 3D meshes, we query GPT-4 regarding
the range of plausible width, height, and depth for each object, and the range of its mass given its  category. We then scale the generated 3D mesh based on the generated size range. 
%3. Inertia parameters depend on mass and . 
We feed the mass and 3D mesh information to MeshLab \cite{meshlab} to get the inertia matrix for the asset.  Our prompts for querying GPT for mass and 3D object size can be found on our website.  We wrap the generated mesh information, its semantic name, as well as the physical parameters into URDF files to be loaded into our simulator.

%\textbf{Twin simulation environment generation} We additionally demonstrate that given an image of the robot's workspace, our method is able to create a twin simulation environment, by detecting and segmenting each object in the scene and generating corresponding 3D simulation assets. Such simulation twin environments are highly useful in training expert policies for real-world transfer.

\subsection{Task Generation, Temporal Decomposition and Reward Function Prediction} \label{sec:prompting}

Given either generated assets or assets obtained from publically available datasets, we prompt LLMs to generate meaningful manipulation tasks considering their affordances, to decompose these tasks into subtasks when possible, and to generate reward functions for each subtask. We train reinforcement learning policies for each (sub)task using the generated reward functions, and then chain them together to solve long horizon tasks. %, which would have been impossible without LLMs' decomposition. 
Our LLM prompts %to generate task descriptions, task decompositions and rewards functions 
contain the following sections: 
%supervisioon while the non-LLM 
%. We show the learnt 

%Given the simulation assets and their URDFs, we prompt GPT-4 \cite{openai2023gpt4} to generate meaningful tasks, together with their language descriptions and reward functions. 

%We design chain-of-thought \cite{wei2023chainofthought} prompts with in-context examples to elicit reasoning while generating tasks and reward functions. Moreover, such a prompting style also provides structure to generations, useful for parsing relevant content in downstream programs for automation. The in-context example consists of the following components:  

%\paragraph{Environment Description} 
\noindent \textbf{1. Asset descriptions.} We use combinations of assets we generate using the method of Section \ref{sec:assets}, as well as articulated assets from PartNet Mobility \cite{xiang2020sapien} and GAPartNet dataset \cite{gapartnet}.  We populate our simulation environment with randomly sampled assets. Then, we extract information from the URDF files including link names, joint types and limits using automated scripts. For example, an asset \texttt{microwave} has parts [\texttt{door}, \texttt{handle}, and \texttt{body}], and joint [\texttt{door-joint}] of type \texttt{revolute} with a joint position range $[0, 1]$. We then describe the extracted configurations of the assets to the LLM, as shown below:

\begin{lstlisting}[language=Python]
The environment contains the following assets:
1.  asset_name: "microwave"
    part_cofiguration:
        Part 1: "body"
        Part 2: "door"
            - link_name: "link_0"
            - joint_name: "joint_0"
            - joint_type: "revolute"
            - limit: [0, 1]
        Part 3: "handle"
            - link_name: "handle_0"
            - joint_name: "handlejoint_0"
            - joint_type: "fixed"
2.  asset_name: "cup"
    part_cofiguration:
        Part 1: cup
            - link_name: "base"
            - joint_name: "base_joint"
            - joint_type: "fixed"
\end{lstlisting}

\noindent \textbf{2. Instructions.}  These  include function APIs that can be used by the LLM to query the pose of the robot end-effector, as well as different assets in the given environment:

\begin{lstlisting}[language=Python]
Available APIs from the simulator are:
# returns the pose of the link
get_pose_by_link_name(asset_name, link_name)
# returns the pose of the robot gripper
get_robot_gripper_pose(asset_name, link_name) 
# returns the state of the joint
get_state_by_joint_name(asset_name, joint_name)
# returns the limit of the joint
get_limits_by_joint_name(asset_name, joint_name)

Note:
1. Only use the available APIs from the simulator. 
2. Generate the reward function code snippets in Python.
\end{lstlisting}

\noindent \textbf{3. Examples of task descriptions and decompositions.} These are question-answer pairs that demonstrate task descriptions and their temporal decompositions. %of how tasks can be decomposed into subtasks.
%language questions can be answered for task decomposition and reward generation.  % can be mapped into code. Examples for task decomposition into subtasks 

\begin{lstlisting}[language=Python]
List meaningful manipulation tasks that can be performed 
in this environment. Give subtask decomposition and the 
order of execution to solve the task. Also, provide the 
reward function for each subtask.

The following tasks can be performed in this environment:
1. Open the Microwave Door
2. Close the Microwave Door
3. Pick Cup
4. Place Cup
5. Put the Cup in the Microwave
   This task needs to be decomposed into sub-tasks:
    - Open the Microwave
    - Pick Cup
    - Place the Cup in the Microwave
\end{lstlisting}

\noindent \textbf{4. Examples of reward functions.} These are task to reward function pairs that present demonstrations of how tasks can be translated to reward functions, as shown below: 
%as well as reward functions for each subtask. 
%\begin{equation}
%\label{eq:reward}
%\lambda_d R_{\text {dist}}+\lambda_p R_{\text {pose}}+\lambda_t R_{\text {grasp}}
%\end{equation}
%Note that the following is just an example for the LLM to use as a reference.
\begin{lstlisting}[language=Python]
Task: OpenMicrowaveDoor
Task Description: open the door of the microwave
```
def compute_reward(env):
    # reward function
    door_handle_pose = env.get_pose_by_link_name("microwave", "handle_0")
    gripper_pose = env.get_robot_gripper_pose()
    distance_gripper_to_handle = torch.norm(door_handle_pose - gripper_pose, dim=-1)
    door_state = env.get_state_by_joint_name("microwave", "joint_0")
    cost = distance_gripper_to_handle - door_state
    reward = - cost

    # success condition
    target_door_state = env.get_limits_by_joint_name("microwave", "joint_0")["upper"]
    success = torch.abs(door_state - target_door_state) < 0.1

    return reward, success
```
\end{lstlisting}
For the example above, the  reward function is  comprised of 1) distance between the end-effector and the target part, and 2) distance between the current and the target pose of an articulated asset, link, or joint.

 We provide  our prompts on our website. We show in Section \ref{sec:experiments} that our method can generalize across assets, suggest diverse and plausible tasks, decomposition and reward functions automatically, using a single in-context example in the prompt, without any additional human involvement.
%\vspace{-0.05in}

\subsection{Sequential Reinforcement Learning for Long Horizon Tasks} \label{sec:RL}
%\vspace{-0.05in}
%Following many previous works \cite{chen2022system,geng2023partmanip,pinto2017asymmetric}, 
We train policies using  Proximal Policy Optimization (PPO) \cite{schulman2017proximal} maximizing the generated reward functions for each subtask. We train RL for each generated subtask in temporal order. Once policy training for a subtask converges, we proceed to the next subtask, by sampling the initial state of the end-effector and the  environment close to the terminal states of the previous subtask. 
%The initial state of the gripper and the environment are sampled from the  resulting states of the previous subtask execution. 
This ensures policies can be temporally chained upon training. %Note that while training till convergence does not guarantee successful policy training, since we decompose the high-level task into  fine-grained simpler subtasks, such a heuristic works practically well in our experiments. \
Our policies are trained per environment using privileged information of the simulation state to accelerate exploration. Such learned policies can be used as demonstration data and distilled into vision-language transformer policies, similar to \cite{brohan2022rt, christen2023learning, jaegle2021perceiver}; we leave this for future work.
\begin{figure}[!htb]
    \centering
    \includegraphics[height=5.7cm,keepaspectratio]{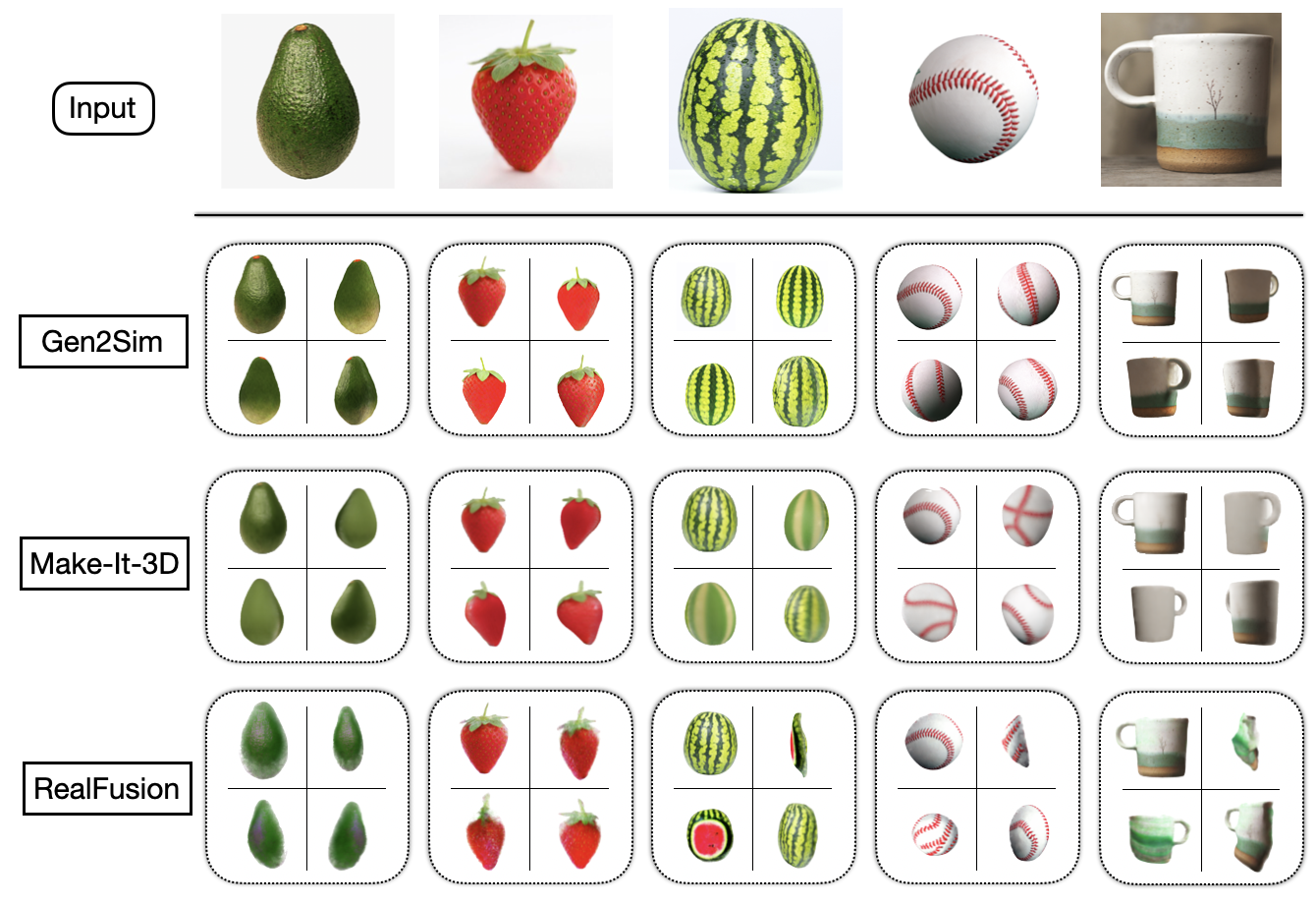}
    \caption{\small\textbf{3D asset generation} from \model{}, RealFusion \cite{melaskyriazi2023realfusion} and Make-It-3D \cite{tang2023makeit3d}. \model{} uses a view and camera pose conditioned image generative model during score distillation,  which helps generate more accurate 3D geometry in comparison to the baselines.} %\textbf{Right:} Our generated simulation objects are shown on top of the table in this simulated scene.}
    \label{fig:asset3dcomparison}
\end{figure}

%\vspace{-0.1in}
\section{Experiments}
\label{sec:experiments}
%\vspace{-0.05in}
%Our experiments aim to 
%We evaluate \model{} in its ability to generate plausible geometry, appearance and physics for a large number and diverse types of objects, in its ability to generate plausible and diverse task goals  and reward function that can drive successful reinforcement learning in simulation. Resulting policies are distilled to vision-based policies to be applied in the real world. 
%In this section, we present comprehensive experiments to answer the following questions:

Our experiments aim to answer the following questions: 

\textbf{1.} Can \model{} generate plausible geometry, appearance, and physics for diverse types of objects and parts, without human expertise and with minimal human involvement? 

\textbf{2.} Can \model{} generate task language goals and reward functions for novel object categories, novel assets with different part configurations, and a combination of multiple assets in an environment?
%Plausibility of our 3D asset generation and URDF creation pipeline by importing assets and building scenes in the simulation environment. 

\textbf{3.} Can the generated environments and reward function lead to successful learning of RL policies?
%in the same way that human developed environments and reward functions do? 
%How do common-sense reasoning from LLMs help with parsing the scene URDF and generating plausible tasks and their reward functions?
%We verify the generated tasks and their corresponding reward function by training policies for k tasks.
% \end{enumerate}
%\vspace{-0.05in}
\subsection{Asset Generation}
\vspace{-0.05in}
We compare our image-to-3D lifting with two baselines: 

1. \textit{RealFusion} \cite{melaskyriazi2023realfusion}, which uses textual inversion of \cite{gal2022image} to learn a word embedding for the depicted object concept in an image, and uses text-conditioned diffusion  with this text embedding  during score distillation.
%adapt a show the plausibility of our pipeline to 

2. \textit{Make-It-3D} \cite{tang2023makeit3d}, which uses the same NeRF and textured mesh two-stage fitting  as \model{}, but does not use a view and pose conditioned generative model, rather a text-based image diffusion model, similar to \cite{poole2022dreamfusion}.

We show comparisons  in Figure \ref{fig:asset3dcomparison}, with images rendered from 4 different views. Our model generates more plausible 3D model as our image diffusion prior comes from an image and pose-conditioned model in comparison to approaches like Fantasia3D or RealFusion which uses text conditioning. 
%Our method is able to generate more plausible 3D assets that show more realistic rendered views compared to the other image-to-3D methods.
We show generated values for 3D sizes and mass for a number of example objects in Table \ref{tab:physics}. We see that  the common sense knowledge encoded in LLMs can produce reasonable physical parameters. %For more visualizations and details of our generated assets, with diversified textures and their behaviors under gravity and collisions, please refer to our website. 

\begin{table}
%\vspace{-1em}
\centering
\small
\tabcolsep=0.1cm
\begin{tabular}{c|cccc} \toprule
                  & Mass (gram) & Length (cm) & Width & Height \\ \midrule
Papaya        & 500-1000            & 15-20                & 10-15   & 10-15 \\
Cucumber   & 200-300               & 15-20                & 5-7      & 5-7 \\
Watermelon & 5000-7000           & 30-40                & 20-30 & 20-30 \\
Raspberry   & 3-5                         & 2-3                    & 2-3      & 2-3 \\
Coconut     & 600-800              & 10-15                 & 8-12    & 8-12 \\
Corn           & 50-100                & 10-15                 & 8-12    & 8-12 \\
Pumpkin    & 2000-5000           & 20-40                & 20-40 & 20-40 \\
Avocado    & 150-250                & 10-12                 & 6-8      & 4-5  \\ \bottomrule
\end{tabular}
\caption{\small\textbf{Size and physics parameter generated by LLMs} for a number of generated assets.} 
\label{tab:physics}
\vspace{-0.2in}
\end{table}

%\begin{figure*}[!htb]
%    \centering
%    \includegraphics[scale=0.25,keepaspectratio]{draft_figures/learning_result.png}
%    \caption{\textbf{Left:} Reward Curves for Trained Policies. \textbf{Right:} Simulation Twin Environments}
%    \label{fig:RL_Twin}
%\end{figure*}

%\vspace{-0.05in}
\subsection{Automated Skill Learning}
%\vspace{-0.05in}
\model{}  generates diverse task descriptions, task decompositions and reward functions automatically for hundreds of assets, with different category labels and number of joints, \textbf{given only a single in-context prompt example} regarding the task decomposition and reward function of the task ``putting a cup in a Microwave'' . Then, the model can generalize to different scenes, asset articulated structures and task temporal lengths. We show some examples of such generated task descriptions  in Figure \ref{fig:teaser} and more on our website. We show examples of  task decompositions in Figure \ref{fig:method}. 
%At the time of submission, our pipeline has generated hundreds of tasks, which we will release upon publication. Note that our method can be queried endlessly to generate more tasks and provide task-specific policy demonstrations, which could be used for policy distillation in the future.
%In this section, we show the ability of LLMs to propose tasks and corresponding reward function that makes sense for a number of diverse assets when appropriately prompted with task and rewards examples.  
We provide our prompts in our project website, alongside examples of the LLM's responses.  

 We learn policies that optimize LLM generated rewards with PPO, an off-the-shelf model-free RL algorithm~\cite{schulman2017proximal}. 
We make use of GPU-parallel data sampling in IsaacGym \cite{makoviychuk2021isaac} for reinforcement learning.  Our robotic setup uses a Franka Panda arm with a mobile base. It is equipped with a parallel-jaw gripper. % or a suction cup, depending on the task needs. %\todo{The suction gripper is only used in pick-and-place tasks where grasping varied geometric objects was intrinsically hard with RL. }
%In all other task categories, we use the parallel jaw gripper. 
Our state representation for PPO includes the robot's joint position $q \in \mathbb{R}^{11}$, velocity $\dot{q} \in \mathbb{R}^{11}$ (7-DoF arm, $x$ and $y$ for the mobile base and 2 extra DoFs from the gripper jaws), orientation of the gripper $r \in S O(3)$, and poses and joint configurations of the assets present in the scene. We use position control and at each timestep $t$ our policy produces target gripper pose and configurations which is  converted to target robot configurations through inverse kinematics. A low-level PID torque controller provided by IsaacGym is  used to produce low-level joint torque commands. We can successfully learn useful manipulation policies, and the polices are able to solve the tasks upon convergence. We show videos of such policies on our website.

\begin{figure}[!htb]
    \centering
    \includegraphics[width=\columnwidth,keepaspectratio]{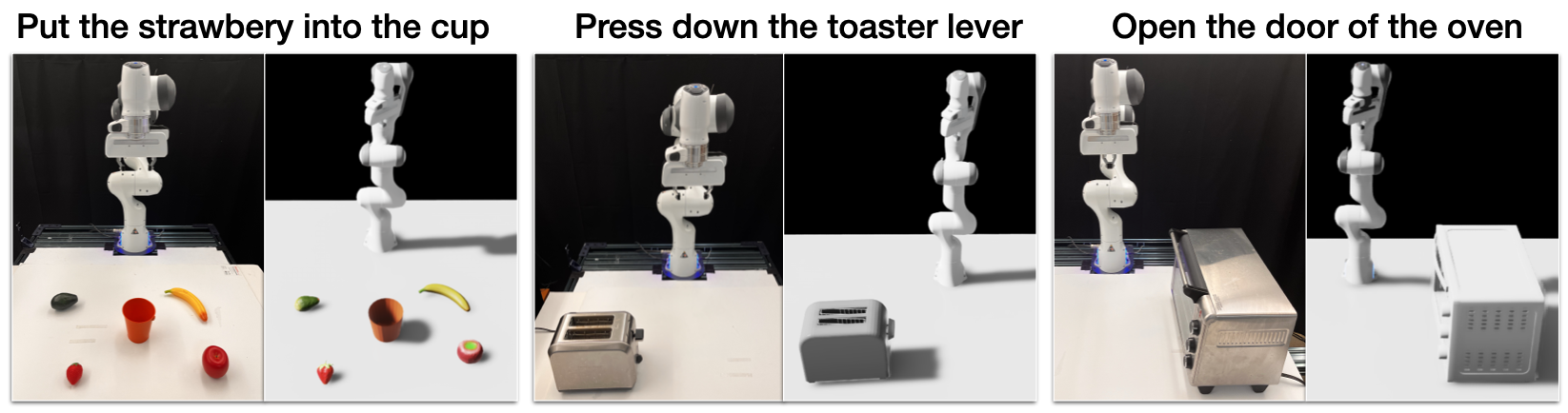}
    \caption{\small\textbf{Twin environments} constructed and generated tasks for sim-to-real transfer. Left: real-world. Right: simulated.} 
    \label{fig:twin}
    \vspace{-0.2in}
\end{figure}

\subsection{Twin environment construction and sim-to-real world transfer}
In order to validate the usefulness of the policies trained in simulation, we construct a twin simulated environment of our lab's real-robot setup (Figure \ref{fig:twin}). We  detect, segment, and estimate the poses of the objects in the scene. For non-articulated assets, we use our model to lift the detected object image to corresponding 3D models; for articulated objects, we select the most similar asset from the \cite{xiang2020sapien}, and populated the simulated environment. %We then apply the estimated pose on these assets and populate the simulation environment with them. 
We train RL policies in simulation and transfer the joint space trajectory back to our real-world setup. Our method allows successful execution of the generated tasks. For more videos of the trained policies and their task executions in simulation, as well as the sim2real transfer, please refer to our website.

% PTAct
%\vspace{-0.1in}
\subsection{Limitations}
%\vspace{-0.05in}
%\todo{our geenrated assets are rigid}
%\todo{we need to mention that RL needs to innovate using subgoals to better handle long horizon}
%Our framework lays out a prototype of how we could scale simulation environment development capitalizing on large-scale pre-trained vision and language generative models. 

\model{} has currently the following two important points to address towards materializing into a platform for large-scale robot skill learning that are deployable in real-world:

%There  are two limitations to addressed for the proposed system to materialize into a platform for large-scale robot skill learning that are deployable in real-world, as identified below:
\textbf{1. Sim2real transfer of closed-loop  policies:} Our current real-world experiments transfer open loop trajectories optimized in the constructed twin environment. For closed-loop policies to transfer to the real world and consume  realistic sensory input, we would need to generate large-scale augmentations for both visual appearances and dynamics for each task and sub-task, and then distil the state-based RL policies to a foundational vision-language policy network. This is a direct avenue for our future work. 

\textbf{2. Beyond rigid asset generation:} The assets we can currently generate are rigid or mostly rigid objects, which do not deform significantly under external forces. For articulated assets, we are using existing manually designed and labelled datasets (\cite{xiang2020sapien, gapartnet}). To generate articulated objects, deformable objects and liquids, accurate fine-grained video perception is required in combination with generative priors to model the temporal dynamics of their geometry and appearance. This is an exciting and challenging direction for future work. 

\section{Conclusion}
\label{sec:conclusion}
%\vspace{-0.05in}
We have presented \model{}, a method for automating the development of simulation environments, tasks and reward functions with pre-trained generative models of vision and language. We presented methods that create and augment geometry, textures and physics of object assets from single images, parse URDF files of assets, generate task descriptions, decompositions and reward python functions, and train reinforcement learning policies to solve the generated long horizon tasks.  
%methods using ground-truth state in the generated simulators. 
Addressing the limitations including generating diverse assets with more complex physical properties, and transfering trained policies to real world are direct avenues for our future work. 
We believe generative models of images and language will play an important role in automating and supersizing robot training data in simulation, and in crossing the sim2real gap, necessary for delivering robot generalists in the real world. \model{} takes one first step in that direction. %towards building large-scale  models for robot control.
%\todo{We hope our work makes a step towards democratizing simulation development so that anyone,  as opposed to resourceful organizations, can develop simulated environments and tasks for robot learning for their  domains of interest.} 

%a language-conditioned manipulation policy architecture that uses recu




 \section*{ACKNOWLEDGMENT}
 This work is supported by Sony AI, NSF award No 1849287, DARPA Machine Common Sense, an Amazon faculty award, and an NSF CAREER award.
% The preferred spelling of the word ÒacknowledgmentÓ in America is without an ÒeÓ after the ÒgÓ. Avoid the stilted expression, ÒOne of us (R. B. G.) thanks . . .Ó  Instead, try ÒR. B. G. thanksÓ. Put sponsor acknowledgments in the unnumbered footnote on the first page.

%%%%%%%%%%%%%%%%%%%%%%%%%%%%%%%%%%%%%%%%%%%%%%%%%%%%%%%%%%%%%%%%%%%%%%%%%%%%%%%%

% References are important to the reader; therefore, each citation must be complete and correct. If at all possible, references should be commonly available publications.

\bibliographystyle{IEEEtran}
\bibliography{references}

\end{document}